\pdfoutput=1

\documentclass[11pt]{article}

\usepackage{acl}

\usepackage{times}
\usepackage{latexsym}
\usepackage{graphicx}
\usepackage{tabularx}
\usepackage{soul}
\usepackage[T1]{fontenc}

\usepackage[utf8]{inputenc}
\usepackage{epstopdf}
\usepackage{hyperref}
\usepackage{xstring}
\usepackage{color}

\usepackage{url}
\usepackage{latexsym}
\usepackage{graphicx}
\usepackage{arydshln}
\usepackage{multirow}
\usepackage{verbatim}
\usepackage{wrapfig}
\usepackage{enumitem}
\usepackage{booktabs}
\usepackage{soul}
\usepackage{tikz}
\usepackage{xcolor}
\usepackage{microtype}
\newcommand\mybox[2][]{\tikz[overlay]\node[fill=blue!10,inner sep=1pt, anchor=text, rectangle, rounded corners=1mm,#1] {#2};\phantom{#2}}

\usepackage{multirow}
\title{How Does Data Corruption Affect Natural Language Understanding Models? A Study on GLUE Datasets}

\author{Aarne Talman\textsuperscript{*}\textsuperscript{$\mathsection$} \hspace{4mm} Marianna Apidianaki\textsuperscript{$\diamond$} \\ \textbf{Stergios Chatzikyriakidis}\textsuperscript{$\dagger$}\textsuperscript{$\ddagger$}\textsuperscript{$\mathsection$} \hspace{4mm} \textbf{J\"org Tiedemann}\textsuperscript{*}\\~\\
\textsuperscript{*}Department of Digital Humanities, University of Helsinki\\
\texttt{\{name.surname\}@helsinki.fi}\\
\textsuperscript{$\diamond$}Department of Computer and Information Science, University of Pennsylvania\\
\texttt{marapi@seas.upenn.edu}\\
\textsuperscript{$\dagger$}Department of Philology, 
University of Crete\\
\texttt{stergios.chatzikyriakidis@uoc.gr}\\
\textsuperscript{$\ddagger$}Centre of Linguistic Theory and Studies in Probability, FLoV, 
University of Gothenburg\\
\textsuperscript{$\mathsection$}Basement AI
}

\begin{document}
\maketitle
\begin{abstract}
A central question in natural language understanding (NLU) research is whether high performance demonstrates the models' strong reasoning capabilities. We present an extensive series of controlled experiments where pre-trained language models are exposed to data that have undergone specific corruption transformations.  These   involve removing instances of specific word classes and often lead to non-sensical sentences. Our results show that performance remains high on most GLUE tasks when the models are fine-tuned or tested on corrupted data, suggesting that they  leverage other cues for prediction even in non-sensical contexts. Our proposed data transformations can be used to assess the extent to which a specific dataset constitutes a proper testbed  for evaluating models' 
language understanding capabilities. 
\end{abstract}

\section{Introduction}

The super-human 
performance of recent Transformer-based pre-trained language models \cite{bert,liu2019roberta} on natural language understanding (NLU) tasks has raised scepticism regarding the quality of the benchmarks used for evaluation \cite{wang-etal-2018-glue,superglue}.  
There is increasing evidence that these datasets contain annotation artefacts and other statistical irregularities that can be 
leveraged by machine learning models to perform the tasks  \cite{gururangan-etal-2018-annotation,poliak-etal-2018-hypothesis,tsuchiya-2018-performance,breakingNLI,talman-chatzikyriakidis-2019-testing,Phametal:2020,talman2021nli}. These studies have so far largely focused on the natural language inference (NLI) and textual entailment tasks.
The scope of our work is wider, in the sense that we address all but one NLU tasks comprised in the GLUE benchmark, specifically: linguistic acceptability (COLA), paraphrasing (MRPC and QQP), sentiment prediction (SST-2), and semantic textual similarity (STS-B). 

\begin{table}[t!]
\small
\centering
\scalebox{0.9}{
        \begin{tabular}{p{.5cm} p{3cm}  p{3cm}} 

 & \parbox{3cm}{\centering \bf Sentence 1} & \parbox{3cm}{\centering \bf  Sentence 2} \\ 
\hline
\multirow{10}{*} {\rotatebox[origin=c]{90}{\parbox{2cm}{\centering paraphrase}}} & \sl \vspace{.5mm}\mybox[fill=blue!10]{\st{Easynews}} \mybox[fill=blue!10]{\st{Inc.}} was subpoenaed late last \mybox[fill=blue!10]{\st{week}} by the \mybox[fill=blue!10]{\st{FBI}}, which was seeking \mybox[fill=blue!10]{\st{account}} \mybox[fill=blue!10]{\st{information}} related to the \mybox[fill=blue!10]{\st{uploading}} of the \mybox[fill=blue!10]{\st{virus}} to the \mybox[fill=blue!10]{\st{ISP's}} \mybox[fill=blue!10]{\st{Usenet}} \mybox[fill=blue!10]{\st{news}} \mybox[fill=blue!10]{\st{group}} \mybox[fill=blue!10]{\st{server}}.\vspace{1.5mm} & \sl \vspace{.5mm} \mybox[fill=blue!10]{\st{Easynews}} \mybox[fill=blue!10]{\st{Inc.}} said \mybox[fill=blue!10]{\st{Monday}} that it was cooperating with the \mybox[fill=blue!10]{\st{FBI}} in trying to locate the \mybox[fill=blue!10]{\st{person}} who uploaded the \mybox[fill=blue!10]{\st{virus}} to a \mybox[fill=blue!10]{\st{Usenet}} \mybox[fill=blue!10]{\st{news}} \mybox[fill=blue!10]{\st{group}} hosted by the \mybox[fill=blue!10]{\st{ISP}}. \\
\hdashline
\multirow{8}{*}{\rotatebox[origin=c]{90}{\parbox{2cm}{\centering non-paraphrase}}} & \sl \vspace{.5mm} \mybox[fill=blue!10]{\st{Arison}} said \mybox[fill=blue!10]{\st{Mann}} may have been one of the \mybox[fill=blue!10]{\st{pioneers}} of the \mybox[fill=blue!10]{\st{world}} \mybox[fill=blue!10]{\st{music}} \mybox[fill=blue!10]{\st{movement}} and he had a deep \mybox[fill=blue!10]{\st{love}} of Brazilian \mybox[fill=blue!10]{\st{music}}. \vspace{1.5mm} & \sl \vspace{.5mm} \mybox[fill=blue!10]{\st{Arison}} said \mybox[fill=blue!10]{\st{Mann}} was a \mybox[fill=blue!10]{\st{pioneer}} of the \mybox[fill=blue!10]{\st{world}} \mybox[fill=blue!10]{\st{music}} \mybox[fill=blue!10]{\st{movement}} -- well before the \mybox[fill=blue!10]{\st{term}} was coined -- and he had a deep \mybox[fill=blue!10]{\st{love}}  of Brazilian \mybox[fill=blue!10]{\st{music}}.\\ 

            \hline
        \end{tabular}
        }
    \caption{\label{table:example_train_data} 
    Example sentence pairs from 
    the corrupted MRPC training dataset where all instances of nouns have been removed. 
    } 
\end{table} 

We present 
a series of experiments where the datasets used for model training and evaluation undergo a number of corruption transformations, which involve removing specific word classes from the data. We remove words pertaining to a specific class (e.g., nouns, verbs), instead of random words, to see the relative importance of word classes for the NLU tasks. For instance, verbs arguably play a significant role in sentence level semantics and removing them is expected to have a bigger impact on the GLUE scores, compared to say determiners. 

 \begin{table*}[t!]
 \small
   \begin{center}
        \begin{tabular}{l l l l}
 &\bf Task  & \bf Baseline &\bf Metric\\
\hline
{\sc cola} & The Corpus of Linguistic Acceptability \cite{warstadt2018neural} & 64.05 & Matthew's correlation \\
{\sc mnli-m} & Multi-Genre Natural Language Inference \cite{multinli} & 87.89 & accuracy \\
{\sc mrpc} & Microsoft Research Paraphrase Corpus \cite{dolan-brockett-2005-automatically} & 88.73 & accuracy \\
{\sc qnli} & Question Natural Language Inference \cite{rajpurkar-etal-2016-squad} & 92.64 & accuracy \\
{\sc qqp} & Quora Question Pairs & 91.32 & accuracy \\
{\sc rte} & Recognizing Textual Entailment \cite{dagan2006} & 70.04 & accuracy \\
{\sc sst-2} & The Stanford Sentiment Treebank \cite{socher-etal-2013-recursive} & 94.61 & accuracy \\
{\sc sts-b} & Semantic Textual Similarity Benchmark \cite{cer-etal-2017-semeval} & 90.08 & Pearson correlation \\
\hline
        \end{tabular}
          \end{center}
    \caption{\label{table:baseline} 
    Baseline results obtained for different GLUE tasks with RoBERTa-{\tt base} and the relevant metric.} 
\end{table*}

The transformations seriously affect the quality of the sentences found in the datasets, making them in many cases unintelligible (cf. examples in Table \ref{table:example_train_data}); a decrease in  performance for models fine-tuned on these corrupted datasets would, thus, be expected. High performance would, instead, indicate that the models rely on lexical cues that remain after corruption, and possibly on other dataset artefacts, to perform a task without necessarily understanding the meaning of the processed utterances. 

Our results show that performance after the corruptions remains high for most GLUE tasks, suggesting that the models leverage other cues for prediction even in non-sensical contexts.

\section{Related Work}
\label{related_work}

Annotation artefacts and statistical biases in NLI datasets are easily leveraged by the models and can guide prediction 
\cite{lai-hockenmaier-2014-illinois,marelli-etal-2014-sick,poliaketal2018,gururangan-etal-2018-annotation}. Examples include explicit negation being indicative of contradiction, and generic nouns  suggesting 
entailment. Artefacts are also present in other types of datasets, for example in the ROC Story dataset 
where models can provide story endings without looking at the actual stories \cite{schwartzetal2017,cai2017}. 
Several works have proposed more challenging and cleaner NLI datasets where artefacts have been removed \cite{mccoy-etal-2019}. An efficient way to do this is using adversarial filtering \cite{nieetal2020,swag}. The superior quality of the  resulting NLI datasets is confirmed by \citet{talman2021nli} in a series of experiments 
where it is shown that data corruption affects these higher quality datasets to a greater extent than previous datasets. 

This work follows the same experimental direction where text perturbations serve to explore the sensitivity of language models to specific phenomena 
\cite{futrell-etal-2019-neural,Ettinger:2020,taktasheva2021shaking,dankers2021}. It has been shown, for example, that shuffling word order causes significant performance drops on a wide range of QA tasks \cite{Sietal:2019,Sugawara:2019}, but that state-of-the-art NLU models are not sensitive to word order \cite{Phametal:2020,sinha2021masked}. Syntax-based perturbations have also been studied in relation to robustness and faithfulness of machine translation models \cite{parthasarathi-etal-2021-sometimes-want}. 

We add to this line of research by applying data corruption transformations that involve removing entire word classes \cite{talman2021nli} to all but one GLUE tasks.\footnote{We exclude WNLI as its development dataset was designed to be adversarial \cite{wang-etal-2018-glue} and hence the corruptions do not have any impact on this dataset when evaluating with the development set.} 
We interpret high performance of models fine-tuned and/or tested on corrupted datasets as an indication of the presence of lexical cues, and possibly artefacts, guiding prediction, 
since the meaning of the corrupted utterances is often hard to recover. 

 \begin{table*}[ht!]
\small
\centering
\scalebox{0.85}{
        \begin{tabular}{l c c | c c | c c}

\bf Data &  \bf {\sc Corrupt-Train} 
&\bf $\Delta$ & \bf {\sc Corrupt-Test}  
&\bf $\Delta$ & \bf {\sc Corrupt-Train and Test} 
&\bf $\Delta$\\
            \hline
{\sc cola-noun} & 39.72 & -24.34 & 17.75 & -46.30 & 34.33 & -29.73 \\
{\sc mnli-m-noun} & 85.64 & -2.24 & 72.85 & -15.04 & 77.46 & -10.42 \\
{\sc mrpc-noun} & 86.27 & -2.45 & 82.35 & -6.37 & 80.15 & -8.58 \\
{\sc qnli-noun} & 89.13 & -3.51 & 71.02 & -21.62 & 82.02 & -10.62 \\
{\sc qqp-noun} & 86.69 & -4.63 & 72.57 & -18.75 & 84.17 & -7.16 \\
{\sc rte-noun} & 47.29 & -22.74 & 53.79 & -16.25 & 47.29 & -22.74 \\
{\sc sst-2-noun} & 94.04 & -0.57 & 87.27 & -7.34 & 88.76 & -5.85 \\
{\sc sts-b-noun} & 81.67 & -8.41 & 56.12 & -33.96 & 63.52 & -26.56 \\
\hdashline
{\sc cola-verb}  & 23.26 & -40.79 & 4.30 & -59.75 & 20.22 & -43.83 \\
{\sc mnli-m-verb}  & 86.95 & -0.94 & 77.61 & -10.28 & 80.32 & -7.57 \\
{\sc mrpc-verb}  & 85.54 & -3.19 & 85.54 & -3.19 & 85.05 & -3.68 \\
{\sc qnli-verb}  & 92.00 & -0.64 & 87.41 & -5.24 & 90.15 & -2.49 \\
{\sc qqp-verb}  & 89.49 & -1.84 & 86.01 & -5.31 & 89.05 & -2.27 \\
{\sc rte-verb}  & 65.34 & -4.69 & 65.70 & -4.33 & 65.34 & -4.69 \\
{\sc sst-2-verb}  & 93.69 & -0.92 & 89.33 & -5.28 & 89.56 & -5.05 \\
{\sc sts-b-verb} & 87.63 & -2.46 & 85.54 & -4.54 & 86.22 & -3.86 \\
            \hline
        \end{tabular}
        }
    \caption{\label{table:results-noun} Example results for the RoBERTa-{\tt base} model 
    fine-tuned on 
    {\sc Corrupt-Train} 
    and tested on the original evaluation set (columns 2 and 3);  fine-tuned on the original data and tested on {\sc Corrupt-Test}; 
    fine-tuned on {\sc Corrupt-Train}  and tested on {\sc Corrupt-Test} 
    (columns 6 and 7). 
    $\Delta$ is the difference to the baseline scores obtained by RoBERTa-{\tt base} on the original dataset, given in Table \ref{table:baseline}.
    }
\end{table*}

\section{Datasets and Corruptions}
\label{datasets}
    
In our experiments, we address eight 
tasks included in 
the General Language Understanding Evaluation (GLUE) benchmark for the English language \cite{wang-etal-2018-glue}: CoLa, MNLI, MRPC, QNLI, QQP, RTE, SST-2, STS-B.  
Following \citet{talman2021nli}, we corrupt the training and development sets available for these tasks by removing words of specific word classes.\footnote{We annotate the original texts with universal part of speech (POS) tags using the NLTK library (\url{https://www.nltk.org/}) and the averaged perceptron tagger.}  We use the development sets for  evaluation, since  
annotated test data have not been made publicly available.
\footnote{For MNLI, we use the matched 
development set \cite{multinli}.} 
We create three configurations for each task: 
(a) {\sc Corrupt-Train}: 
fine-tuning on the corrupted training set, 
evaluation on the original development set; (b) {\sc Corrupt-Test}: fine-tuning 
on the original training set, evaluation on 
the corrupted test set; (c) {\sc Corrupt-Train and Test}: training and evaluation 
on corrupted data. The corruption procedure involves removing all instances of a specific word class from the corresponding dataset ({\sc adj, adv, conj, det, noun, num, pron, verb}). We label the corrupted datasets by indicating the class of the words that have  
been removed  
(e.g., {\sc cola-noun}, {\sc qnli-verb}). Given  
the possible combinations of tasks, datasets and corruptions, 
we end up with 192 setups for 
our experiments.

Note that the resulting sentence fragments do not constitute propositions. Although not ideal, this is not necessarily problematic for tasks such as sentiment analysis. For inference, the assumption that the task can only be performed at the propositional level is a strong claim, especially given that examples which are not propositions are abundant in existing benchmarks such as MNLI (e.g., examples extracted from dialogue).

\section{Models}
\label{evaluation}
 
 We fine-tune 
 the pre-trained RoBERTa-{\tt base} model \cite{liu2019roberta} from the Huggingface Transformers library \cite{wolf-etal-2020-transformers} 
 in  each of our 192  configurations. We use 
the same fine-tuning and evaluation set up for all the experiments.  We retrieve the GLUE datasets using the Huggingface Datasets library \cite{2020HuggingFace-datasets}.  We fine-tune the models for 3 epochs, using a batch size of 32 and a learning rate of 0.00002.

 \section{Results}
 \label{section:results}
 
\begin{figure}[t!]
   \scalebox{0.85}{
\includegraphics[width=7.5cm]{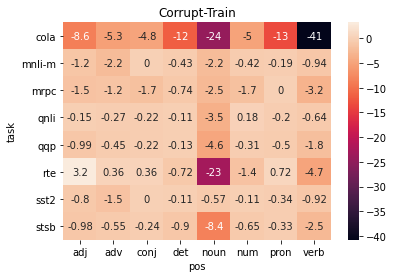}}
  \caption{Impact of specific 
  data corruptions in the {\sc Corrupt-Train} setting. The columns correspond 
  to the removed word class 
  and the rows to the GLUE tasks.}
    \label{fig:Corrupt-Train}
\end{figure}

\begin{figure}[h!]
    \scalebox{0.85}{
\includegraphics[width=7.5cm]{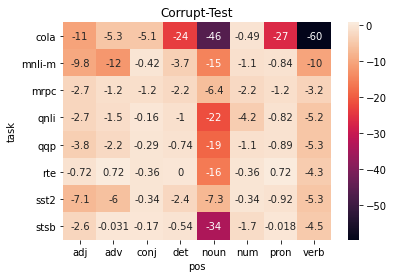}}
  \caption{Impact of specific data corruptions in the {\sc Corrupt-Test} setting for each task.}  
    \label{fig:Corrupt-Test}
\end{figure}

\begin{figure}[h!]
   \scalebox{0.85}{
\includegraphics[width=7.5cm]{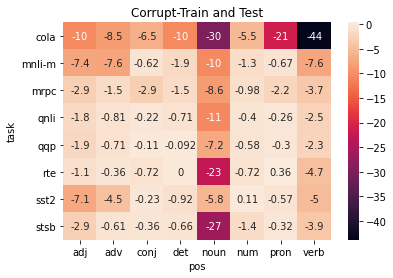}}
  \caption{Impact of specific data corruptions in the {\sc Corrupt-Train and Test} setting for each task. 
  }
    \label{fig:Corrupt-Train_and_Test}
\end{figure}

 The baseline results using the original (non-corrupted) 
 datasets  are shown in Table \ref{table:baseline}. Given the large number of ~configurations, 
 ~we only report the exact evaluation results for the -{\sc noun} and  {-{\sc verb}} settings in Table  \ref{table:results-noun}, 
 as these content word classes arguably contribute a lot to the meaning of utterances. 
 For the remaining configurations, we visualise the effect of the corruptions using heatmaps that show the difference in performance compared to the baseline results  (Figures \ref{fig:Corrupt-Train} to 
 \ref{fig:Corrupt-Train_and_Test}).

\begin{table*}[h!]
 \small
 \scalebox{0.99}{
 \begin{tabular}{l | l | l  | l | l}
 \bf \parbox{3.8cm}{\centering Original Sentences \vspace{1mm}} & \parbox{3.2cm}{\centering {\sc Corrupt-Test-Noun} \vspace{1mm}} &  \bf \parbox{3.2cm}{\centering {\sc Corrupt-Test-Adj}  \vspace{1mm}} &  \bf \parbox{8mm}{\centering Labels} & \bf \parbox{1.4cm}{\centering Gold label} \\
\hline
\parbox{4.2cm}{\vspace{1mm} \sl An unclassifiably awful study in self - and audience-abuse.} & \parbox{3.2cm}{\vspace{1mm}\sl an unclassifiably awful in - and. \vspace{1mm}} & \parbox{3.2cm}{\vspace{1mm}\sl an unclassifiably study in self - and audience-abuse.} & positive &\parbox{1.4cm}{\vspace{1mm} \centering {negative} } \\ \hline 
\parbox{4.2cm}{\vspace{1mm}\sl It proves quite compelling as an intense, brooding character study.} & \parbox{3.2cm}{\vspace{1mm}\sl it proves quite compelling as an intense, brooding.}  & \parbox{3.2cm}{\vspace{1mm}\sl it proves quite as an, brooding character study. } & positive & \parbox{1.4cm}{\vspace{1mm} \centering {positive} }\\ 
        \end{tabular}
        }
    \caption{\label{table:NRCLexPredictions} Labels assigned by NRCLex to sentences from the {\sc sst-2} {\sc Corrupt-Test-Noun/-Adj} datasets. 
    }
\end{table*}
 
 Our results for the {\sc -noun} and {\sc -verb} corruptions in {\sc Corrupt-Train} (Table \ref{table:results-noun}), and for all configurations in Figure \ref{fig:Corrupt-Train}, show a notable decrease in performance on {\sc cola} and {\sc rte,}  especially when nouns are removed. 
The impact on {\sc mnli-m} and {\sc qnli} datasets is small, confirming previous findings regarding the presence of annotation artefacts and lexical cues that can guide model prediction. Our results suggest that this is the case also in other GLUE datasets, such as {\sc mrpc} and {\sc sst-2}, where the models still manage to perform fairly  well compared to the baseline when fine-tuned on corrupted data. 

Our {\sc Corrupt-Test} results in Table  \ref{table:results-noun} and in Figure \ref{fig:Corrupt-Test} 
show that removing nouns from the data used for evaluation  has a  much larger impact across tasks, compared to {\sc Corrupt-Train}. The 
 biggest drop in performance is observed  on {\sc cola}, {\sc mnli-m} and {\sc sts-b}. However, accuracy on 
 {\sc mrpc} and {\sc sst-2} is 
 still very high, suggesting 
 that good 
 performance  does not  require sentence-level understanding but can be achieved by relying on 
 lexical cues present in the data. In the {\sc Corrupt-Train and Test} setting (Table \ref{table:results-noun} and  Figure \ref{fig:Corrupt-Train_and_Test}),  we observe the biggest  drop in performance  on {\sc cola}, {\sc mnli-m} and {\sc sts-b}, and a  lower 
 impact on {\sc qnli}, {\sc qqp} and {\sc sst-2}.


\begin{table}[h!]
  \begin{center}
  \scalebox{0.8}{
  \begin{tabular}{p{2cm} p{3cm}  p{2cm}} 
\bf Word class & \bf Dataset & \bf Accuracy \\
\hline
{\sc Noun} & {\sc Corrupt-Test} & 14.7\% \\
{\sc Noun} & original & 34.1\% \\
\hdashline
{\sc Verb} & {\sc Corrupt-Test} & 31.1\%\\
{\sc Verb} & original & 66.4\%\\
\hline
        \end{tabular}
}
 \end{center}
    \caption{\label{table:masked_LM} Accuracy of RoBERTa{\sc-base} in predicting a masked word in the {\sc mrpc} development set.} 
\end{table}

\section{Discussion and Analysis}
\label{Discussion}
\subsection{Lexical Cues}
Our results show that model performance in many tasks is marginally affected by the imposed corruptions which, however, in many cases alter the meaning of utterances. 
We conduct additional analyses 
aimed at identifying the lexical cues that remain after corruption 
and can guide model prediction. We focus on {\sc mrpc} (Microsoft Research Paraphrase Corpus) and {\sc sst-2} (Stanford Sentiment Treebank), where the impact of {\sc Corrupt-test} transformations was the smallest.

{\sc mrpc} addresses the paraphrase relationship between sentence pairs. We explore the semantic similarity of the information that remains after corruption. Our assumption is that if a sentence pair (from which nouns or verbs have been removed) still contains synonyms or longer paraphrases, this can guide the model towards detecting a similarity or entailment relationship. For this analysis, we use the unigram paraphrases in the L (large) package of  
PPDB 2.0 \cite{pavlick-etal-2015-ppdb}. 
We find that in the {\sc Corrupt-Test-Noun} {\sc mrpc} dataset, 
76\% of the sentence pairs for which the model made correct predictions still include a lexical paraphrase. 

{\sc sst-2} involves detecting the 
sentiment expressed in individual sentences. 
We use the NRCLex tool\footnote{NRCLex is based on the expanded version of the NRC Word-Emotion Association Lexicon \cite{mohammad-turney-2010-emotions,Mohammad2013CROWDSOURCINGAW}. 
We only use the `positive' and `negative' keys.
}  to measure the sentiment 
expressed by lexical cues in the {\sc Corrupt-test} sentences for which 
model predictions are correct.  
Given that sentiment can be expressed in a text by words pertaining to different grammatical categories, 
we explore whether lexical cues indicating the polarity 
of the text 
still remain after removing instances of a specific word class. 
In Table \ref{table:NRCLexPredictions}, we show the labels predicted by NRCLex for corrupted test sentences, where the nouns and adjectives have been dropped. 
We observe that even if sentences become non-sensical after corruption, 
it is  still possible to detect the (positive or negative) polarity of the sentences  from the remaining words. Relying on these lexical cues, RoBERTa often manages to predict 
the correct sentiment. Specifically, according to the  NRCLex predictions, the correct sentiment is still present in 383 out of 761 corrupted sentences where RoBERTa made correct predictions in the {\sc Corrupt-Test-Noun} setting. If both nouns and adjectives are removed 
({\sc Corrupt-Test-Noun-Adj}), 
NRCLex detects that the correct sentiment is still present in 125 out of the 672 examples that were correctly predicted by RoBERTa.


\subsection{Can RoBERTa Guess the Missing Tokens?} 

As RoBERTa has been pre-trained using a Masked Word Prediction task, it is reasonable to ask if high model performance with our corrupted datasets could be due to the model's ability to ``fill in the gaps'' and predict the missing words. To test this, in each sentence of the {\sc mrcp} development set,  we replace the first token that is  
aimed by a specific corruption procedure (-{\sc noun/verb}) 
with the [MASK] 
token. We do this in the original sentence (by removing only the first noun/verb instance) and in the corrupted sentence (where all other nouns/verbs are missing). 
For example, from the first sentence  in Table \ref{table:NRCLexPredictions}, we generate two cloze-task queries in the -NOUN setting: 
\begin{enumerate}[label=(\alph*),noitemsep]
    \item An unclassifiably awful [MASK] in self - and audience-abuse.
    \item An unclassifiably awful [MASK] in - and.
\end{enumerate}

We use these queries to test RoBERTa's token prediction capability. As shown in Table \ref{table:masked_LM}, 
it is easier   
to predict the masked token  
in the original 
sentences, but the model 
is still able to make correct predictions in the corrupted sentences. 
This could partly explain the  
high performance observed 
for  {\sc mrpc} in the corrupted setting (cf. Section \ref{section:results}).

\section{Conclusion}
\label{conclusion}
We apply a set of controllable corruption transformations to the datasets of  
 NLU tasks in the GLUE benchmark, and study their impact on model performance.  
The proposed transformations 
are generic enough to be applicable to other NLU tasks, and can enrich the available artillery for 
dataset quality assessment in terms of how efficiently they trigger and test the language understanding capabilities of the models. Our results indicate that understanding  the meaning of utterances is not required for high performance in most GLUE 
tasks. 
This finding suggests caution in interpreting leaderboard results and in the conclusions that can be drawn regarding the language understanding capabilities of the models. We make our code  available\footnote{\url{https://github.com/Helsinki-NLP/nlu-dataset-diagnostics}}
in order to promote the application of these tests 
to other NLU datasets, and to  favour the development of benchmarks addressing the actual  capability of the models  
to reason about language. 


\section*{Acknowledgments}
\vspace{1ex}
\noindent

\begin{wrapfigure}[]{l}{0pt}
\includegraphics[scale=0.3]{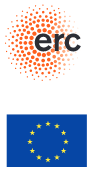}
\end{wrapfigure}

\noindent During the course of this project, Marianna Apidianaki and Jörg Tiedemann were supported by the FoTran project, funded by the European Research Council (ERC) under the European Union’s Horizon 2020 research and innovation programme (grant agreement no.~771113). Stergios Chatzikyriakidis is supported by grant 2014-39 from the Swedish Research Council, which funds the Centre for Linguistic Theory and Studies in Probability (CLASP) in the Department of Philosophy, Linguistics, and Theory of Science at the University of Gothenburg. We thank the reviewers for their thoughtful comments and valuable suggestions.

\bibliography{anthology,custom}




\end{document}